# NP-ODE: Neural Process Aided Ordinary Differential Equations for Uncertainty Quantification of Finite Element Analysis

Yinan Wang[1], Kaiwen Wang[2], Wenjun Cai[2*], Xiaowei Yue[1*]

[1]*Grado Department of Industrial and Systems Engineering, Virginia Polytechnic Institute and State University, Blacksburg, USA*

[2]*Materials Science and Engineering, Virginia Polytechnic Institute and State University, Blacksburg, USA*

[*]corresponding author: Dr. Wenjun Cai, Dr. Xiaowei Yue. Email: caiw@vt.edu, xwy@vt.edu

## Abstract

Finite element analysis (FEA) has been widely used to generate simulations of complex and nonlinear systems. Despite its strength and accuracy, the limitations of FEA can be summarized into two aspects: a) running high-fidelity FEA often requires significant computational cost and consumes a large amount of time; b) FEA is a deterministic method which is insufficient for uncertainty quantification (UQ) when modeling complex systems with various types of uncertainties. In this paper, an physics-informed data-driven surrogate model, named Neural Process Aided Ordinary Differential Equation (NP-ODE), is proposed to model the FEA simulations and capture both input and output uncertainties. To validate the advantages of the proposed NP-ODE, we conduct experiments on both the simulation data generated from a given ordinary differential equation and the data collected from a real FEA platform for tribocorrosion. The performances of the proposed NP-ODE and several benchmark methods are compared. The results show that the proposed NP-ODE outperforms benchmark methods. The NP-ODE method realizes the smallest predictive error as well as generates the most reasonable confidence interval having the best coverage on testing data points.

# 1. Introduction

## 1.1 Motivation

Finite element analysis (FEA) is a powerful tool to simulate physical phenomena or operations, which is a numerical method for solving differential equations. Typically, to analyze a large and complex system, the FEA method will first divide the entire system into small pieces called finite elements, which is achieved by constructing the meshing of the object. The equations that describe each finite element will then be assembled to model the entire system. The approximated solution of the entire system can be finally generated by solving differential equations of all finite elements under environments with a combination of loads and constraints. Because FEA provides outstanding accuracy when modeling complex nonlinear systems, it has been applied to many fields, which include advanced manufacturing, new material design, heat transfer, solid/fluid mechanics, and multiphysics systems (Zienkiewicz et al., 2014).

Despite the strength of the FEA method in accurately analyzing complex systems, the application of FEA is hindered by several limitations, which can be summarized into two folds. On the one hand, the high-fidelity FEA simulation usually requires a high computational cost (Su et al., 2017, Wang et al., 2020). This issue will be extraordinarily critical in some scenarios, such as using ultra-fine elements in FEA, repeating simulations for the function validation, etc. On the other hand, FEA is a deterministic simulation without explicitly considering the system uncertainties. For example, intrinsic uncertainty can come from input measurement errors or parameter setting deviations, and extrinsic uncertainty may be caused by computational and measurement errors for responses. In most systems with complicated nonlinear behaviors, it is challenging to obtain deterministic and accurate measurements of the parameters when building up the FEA so that uncertainty quantification is highly demanded in FEA (Mahadevan et al., 2011, Wang et al., 2021).

## 1.2 Literature Review

Recent literature has proposed many methods to tackle the aforementioned limitations of the FEA method, which can be divided into two branches. One branch of research deals with high computational costs by using surrogate models to approximate and replace the FEA method in system modeling. The other branch of literature takes the system uncertainties into modeling by either directly designing stochastic simulation methods or building stochastic surrogate models to replace the FEA method. A detailed review of these two branches of works is introduced as follows.

*1.2.1 Review on Deterministic Surrogates*

Data-driven deterministic surrogates are commonly used to approximate and replace the FEA method. The basic idea is to train and validate the data-driven surrogate model based on the inputs and outputs of FEA, where uncertainties in the simulation output given the same input are ignored. As long as the trained surrogate is accurate, it can bypass the FEA calculation process and replace FEA in further applications. Kriging or Gaussian Process (GP) regression is widely used as surrogates. Yue and Shi (2018) proposed a surrogate model-based optimal feed-forward control strategy by using a universal kriging model for dimensional variation reduction and defect prevention in the assembly process of composite structures. Wang et al. (2020) proposed a GP constrained general path model to approximate the high-fidelity FEA for efficient product and process design in additive manufacturing. Deep Neural Networks (DNN) are another type of surrogates. Dong et al. (2020) proposed a systematic implementation of Artificial Neural Network (ANN) in replacing FEA in the structural optimization of elastic metamaterials. Liang et al. (2018) developed a deep learning (DL) model to replace structural FEA in estimating the stress distributions of the aorta. Although the proposed data-driven surrogates provide an efficient alternative to the FEA method, a common limitation among these pure data-driven models was the lack of physical insights, which limited the application of the resulting models.

Considering the limitations of pure data-driven surrogates, in the domain of machine learning, physics-informed data-driven surrogates have been proposed, which incorporated the physical insights into surrogate design. Loose et al. (2009) derived one physical-analysis driven surrogate model base on first principles. Considering most of the physical phenomena can be modeled by differential equations, applying Neural Network in solving differential equations has been researched. Chen et al. (2018) proposed Neural Ordinary Differential Equations (Neural-ODE), which applied neural networks to parameterize the derivatives of underlying function between the input and the output instead of modeling their mapping directly. Yildiz et al. (2019) proposed a latent second-order ODE model for high-dimensional sequential data, which explicitly decomposed the latent space into momentum and position components and analyzed a second-order ODE system. The Neural-ODE showed its strength in analyzing systems that are governed by differential equations, which may have a great potential to be a suitable surrogate model of FEA simulations.

*1.2.2 Review on Uncertainty Quantification and Stochastic Surrogates*

Researchers incorporated system uncertainties into designing the simulation method or proposing stochastic surrogates. Uncertainties are inevitably existing in real systems, which may be caused by sensing errors, actuating errors, computational errors, etc. FEA simulation is a deterministic method that is insufficient in capturing system uncertainties. To tackle this issue, the Monte Carlo simulation is widely used to capture the uncertainties in the simulation of complex nonlinear systems (Rubinstein and Kroese, 2017). Intuitively, this method will repeat simulations multiple times assuming different values of unknown parameters, which makes the computational cost grows significantly when the system has a large number of unknown parameters.

Similarly, stochastic surrogates have been proposed to consider the system uncertainties as well as reduce the computational cost (Wang and Shan, 2006). One body of stochastic surrogates is GP-based models. Ankenman et al. (2010) extended the kriging to stochastic simulation setting and proposed stochastic kriging for simulation metamodeling, which characterized both the intrinsic uncertainty inherent in stochastic simulation and the extrinsic uncertainty about the unknown response surface. A surrogate model with considering diverse uncertainty sources was proposed to achieve better predictive assembly of composite aircrafts (Yue et al., 2018). There are some GP-based methods specifically considering intrinsic (input) uncertainty. Wang et al. (2019) refined the GP-based optimization algorithms to solve the stochastic simulation optimization problems considering input uncertainty. Wang et al. (2020) investigated GP regression considering the input location error within FEA simulations. GP-based methods have shown their advantages as the stochastic surrogate, but their applications are hindered by two limitations: 1) the choice of covariance kernel will significantly influence the model performance; 2) they are computationally prohibitive for large and high dimensional datasets (Snelson and Ghahramani, 2006). Hoang et al. (2015) proposed the stochastic variational GP (SVGP) to train a sparse GP at any time with a small subset of training data, which improves the scalability of GP-based methods in dealing with "Big Data." However, the influence of approximated representation in sparse GP on model accuracy can be nonnegligible, and the parameters introduced by variational inference require extra training time (Liu et al., 2020).

Another body of stochastic surrogate is the extension of DNN. Blundell et al. (2015) proposed a Bayesian Neural Network (BNN) to enable NN to model uncertainty by learning a distribution of model weights instead of specific values. Gal and Ghahramani (2016) proposed Neural Network (NN) with dropout to capture the variations in training data and evaluate the uncertainties in new simulations accordingly. The NN-based stochastic surrogates showed their strength in modeling large datasets. However, the model performance is highly related to the choice of hyperparameters, such as the prior distribution of model weights in BNN and the dropout ratio in NN with dropout. Furthermore, learning posterior distributions in BNN can be difficult considering the high dimensionality and complexity.

Inspired by the idea of GP, Garnelo et al. (2018) proposed a family of models Conditional Neural Processes (CNPs) that combine neural networks with features of GP. This method is to define conditional distributions over possible functions given a set of observations. In this way, the CNPs could generate predictive results as well as evaluating the uncertainties of output. Next, a more general model, Neural Processes (NPs), was proposed to add a latent encoding branch to project the input uncertainties into global latent representations and then incorporate latent representations into the generation process of output distributions (Garnelo et al., 2018). Kim et al. (2019) incorporated the attention module (Bahdanau et al., 2014) into NPs to propose Attentive Neural Processes (ANPs). The attention module can be regarded as a measure of similarity among context inputs (observed) and target inputs (unobserved) to find which context is most relevant to a given target. Overall, the NPs can characterize the input and output uncertainties. However, these methods cannot capture the differential equations structure of FEA. Intuitively, they are not sufficient for interpretability.

*1.3 Proposed Method and Contributions*

Based on the reviews of recent literature, there is a need to develop a stochastic surrogate with physical insight to reduce the computation cost of the FEA method and capture system uncertainties. To fill in this research gap, we propose a novel method, Neural Process Aided Ordinary Differential Equations (NP-ODE), to build an physics-informed data-driven surrogate for FEA simulations with uncertainty quantification. In our proposed NP-ODE, its structure follows the encoder-decoder format, in which the encoder part is to project the observed simulations into feature space, and the decoder part will take the features and unobserved query as the input to learn distributions over the output. Compared with a pure data-driven decoder, the NP-ODE method incorporates the Neural-ODE as the decoder to strengthen the model in solving complex systems governed by differential equations. To this extent, both the FEA method and our proposed NP-ODE are built to approximate solutions for systems with underlying differential equations. So incorporating Neural-ODE as the decoder makes the model a convincing surrogate for the FEA method. What is more, a similar structure to the NPs can enable uncertainty quantification in our proposed NP-ODE. The pipeline of our proposed methods is shown in Fig. 1. Firstly, an FEA simulation platform is built according to the system's properties and parameters. The FEA platform is further validated by real experiment data and used for generating simulations. Given the generated simulations from FEA, our proposed method NP-ODE is trained and tested on the datasets to generate the predictive output and conduct uncertainty quantification. Finally, the NP-ODE is ready to be used as a stochastic surrogate of FEA for new inputs.

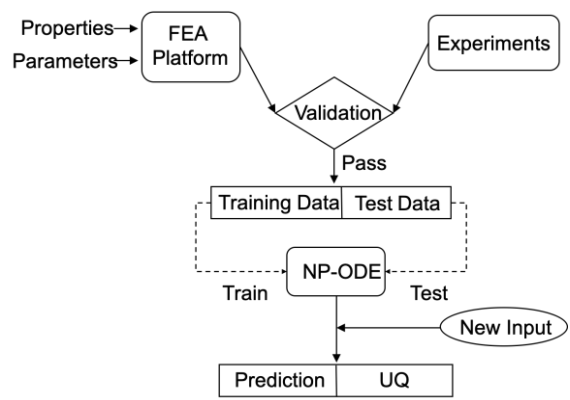

Figure 1. Overview of Building Neural-ODE from FEA Simulations

The contributions of this paper can be summarized into four aspects:

1) the NP-ODE evaluates both the input and output uncertainties and generates distributions over the output to enable uncertainty quantification;

2) compared with the pure data-driven surrogate, the NP-ODE solves differential equations in its decoders, so it is more physically close to the original FEA, and have better interpretability;

3) compared with the original decoder in NPs, NP-ODE can reduce the number of parameters by incorporating Neural-ODE to mitigate the risk of overfitting with scarce training samples;

4) The NP-ODE can improve the robustness in dealing with noisy data points governed by the differential equation.

The remainder of this paper is organized as follows: Section 2 introduces the mathematical foundations of Neural-ODE and NPs; Section 3 formulates the Neural-ODE to model FEA simulations, proposes the NP-ODE method, investigates the uncertainty quantification and properties of the proposed method, and develops a computational algorithm for the proposed NP-ODE; Section 4 presents a simulation study as a proof-of-concept; Section 5 presents the case study, including the introduction of FEA simulation for tribocorrosion, implementation details, evaluation metrics, benchmark models, and performance comparisons between the proposed method and benchmark methods; Section 6 gives a brief conclusion on our proposed method and its advantages.

## 2. Neural ODE and NPs

In this section, we firstly introduce the Neural-ODE and NPs. It is worth mentioning that the notations appeared in this paper are summarized in Appendix I.

### *2.1 Neural Ordinary Differential Equations*

The Neural-ODE was proposed to parameterize the derivative of the hidden state using a neural network (Chen et al. 2018), in which the Neural-ODE showed its strength in applications of image classification, modeling continuous flows, and modeling time-series data.

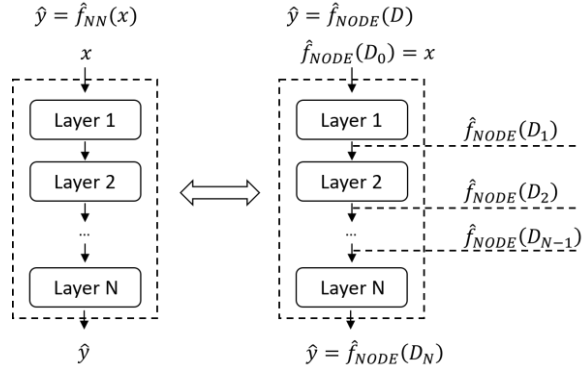

Figure 2. Equivalence Between $\hat{f}_{NN}(x)$ and $\hat{f}_{NODE}(D)$

We take FEA simulation as one example. For simplicity, suppose both the input and output of the FEA method are scalars, and FEA simulations generate $N$ data points $\{(x_1, y_1), \ldots, (x_N, y_N)\}$ with the underlying functional relationship $y = f(x), x \in \mathbb{R}, y \in \mathbb{R}$. The classical neural network is built to approximate the functional relationship by minimizing a loss function $\mathcal{L}\left(\hat{f}_{NN}(x), y\right)$, in which $\hat{f}_{NN}(x)$ is the approximated mapping given by the neural network. As shown in Fig. 2, considering the multi-layer structure of the neural network, $\hat{f}_{NN}(x)$ can be regarded as a series of discrete transformations with input $x$ as the initial value, and the steps of transformations are denoted as the model depth. In this case, the $\hat{f}_{NN}(x)$ is equivalent to $\hat{f}_{NODE}(D)$, $\hat{f}_{NODE}(D_0) = x$, in which $\hat{f}_{NODE}(D)$ represents the discrete transformation with respect to (w.r.t) model depth $D$, and $D_0$ represents the initial layer. Instead of directly modeling the functional relationship as discrete transformations, assuming $\hat{f}_{NODE}(D)$ is continuous and differentiable, the Neural-ODE is built to approximate the first-order derivative of $\hat{f}_{NODE}(D)$. The optimal model depth $D$ will be optimized during the training phase.

$$g_{NODE}(D) = \frac{d\hat{f}_{NODE}(D)}{dD} \tag{1}$$

in which, the $g_{NODE}(D)$ denotes the first-order derivative given by Neural-ODE. Based on the approximated first-order derivative of $\hat{f}_{NODE}(D)$, the output can be calculated by the ODE solver and is shown in equation (2).

$$\hat{f}_{NODE}(D_N) = \text{ODESolve}(\hat{f}_{NODE}(D_0), g_{NODE}(D), D_0, D_N) \tag{2}$$

## 2.2 Neural Processes

The NPs are inspired by the ideas of GP and designed for modeling the regression function, such as $\mathbf{y} = \mathbf{f}(\mathbf{x})$ between input $\mathbf{x} \in \mathbb{R}^m$ and output $\mathbf{y} \in \mathbb{R}^p$ (Kim et al., 2019). Instead of specific values, NPs output distributions over unobserved target FEA data points $(\mathbf{x}_{n+1}, \mathbf{y}_{n+1}), \ldots, (\mathbf{x}_{n+T}, \mathbf{y}_{n+T})$ conditioned on the observed historical FEA simulation data points $(\mathbf{x}_1, \mathbf{y}_1), \ldots, (\mathbf{x}_n, \mathbf{y}_n)$. The model structure of NPs is shown in Fig. 3, in which the output of NPs, $\hat{\mathbf{y}}_{n+1} \sim \mathcal{N}(\bar{\mathbf{y}}_{n+1}, \boldsymbol{\sigma}_{n+1})$, is expressed as the posterior distribution given observed data points as shown in equation (3) (Kim et al., 2019).

$$p(\hat{\mathbf{y}}_{n+1}|\mathbf{x}_{n+1}, (\mathbf{x}_{1:n}, \mathbf{y}_{1:n})) = \int p(\hat{\mathbf{y}}_{n+1}|\mathbf{x}_{n+1}, \mathbf{d}_C, \mathbf{z}) q(\mathbf{z}|\mathbf{s}_C) d\mathbf{z} \quad (3)$$

In equation (3), the $q(\mathbf{z}|\mathbf{s}_C), \mathbf{z} \in \mathbb{R}^N, \mathbf{s}_C \in \mathbb{R}^N$ is the prior distribution on $\mathbf{z}$ given by stochastic encoder, $\mathbf{d}_C \in \mathbb{R}^N$ is given by deterministic encoder, and the likelihood function $p(\hat{\mathbf{y}}_{n+1}|\mathbf{x}_{n+1}, \mathbf{d}_C, \mathbf{z})$ is represented by the decoder. It is worth noting that in FEA simulation, the data points $(\mathbf{x}_{n+1}, \mathbf{y}_{n+1}), \ldots, (\mathbf{x}_{n+T}, \mathbf{y}_{n+T})$ are from different replications and do not have temporal dependencies or a specific order. Equation (3) shows how to predict one unobserved data point $(\mathbf{x}_{n+1}, \hat{\mathbf{y}}_{n+1})$ given observations, which can be directly extended to predict other unobserved data points.

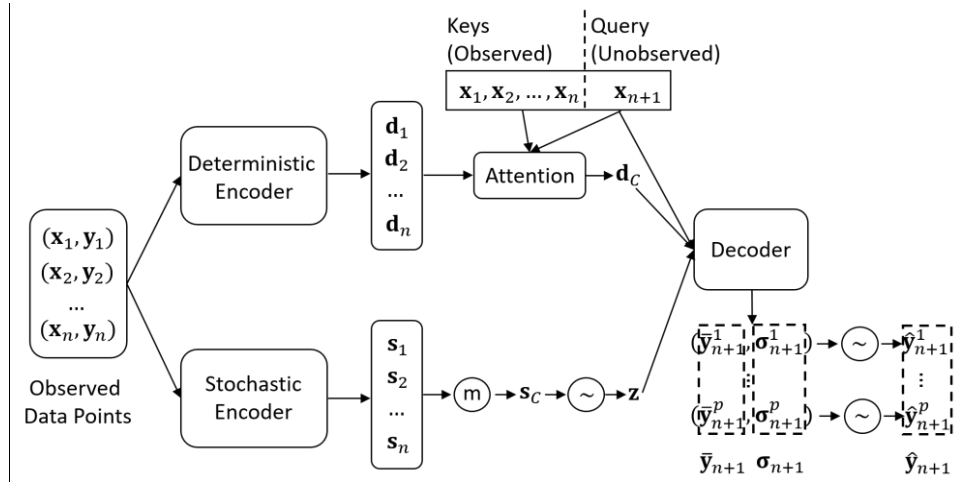

Figure 3. Model Structure of Neural Process

In the encoder part of NPs, the basic building block is the multi-layer perceptron (MLP). On the one hand, the deterministic encoder will capture the interactions among observed FEA data and output the finite-dimensional representation $\mathbf{d}_C$, in which the attention module is an important part to aggregate the representations $(\mathbf{d}_1, \ldots, \mathbf{d}_n)$ from each observed data point corresponding to a specific unobserved target simulation point $\mathbf{x}_{n+1}$. Intuitively, the idea of the attention module is to compute the weights of each observed keys $(\mathbf{x}_1, \ldots, \mathbf{x}_n)$ w.r.t. the unobserved query $\mathbf{x}_{n+1}$, and apply these weights to compute the weighted sum of representations $(\mathbf{d}_1, \ldots, \mathbf{d}_n)$ to generate the output $\mathbf{d}_C$ (Bahdanau et al., 2014). On the other hand, the stochastic encoder will capture the uncertainties in the predictive value $\hat{\mathbf{y}}_{n+1}$ given observed data points and output the prior distribution $q(\mathbf{z}|\mathbf{s}_C)$ on the latent representation $\mathbf{z}$. Because the stochastic representation $\mathbf{s}_C$ of the context data points are generated by mean aggregation of $(\mathbf{s}_1, \ldots, \mathbf{s}_n)$, the prior distribution $q(\mathbf{z}|\mathbf{s}_C)$ is invariant to permutations of context data points. The role of the latent representation $\mathbf{z}$ is to account for the uncertainties in predicting $\hat{\mathbf{y}}_{n+1}$ given observed data. The decoder will take the unobserved query $\mathbf{x}_{n+1}$, deterministic representation $\mathbf{d}_C$ of observed data w.r.t. $\mathbf{x}_{n+1}$, and stochastic representation $\mathbf{s}_C$ of observed data as the input, output the mean $\bar{\mathbf{y}}_{n+1}$ and standard deviation $\boldsymbol{\sigma}_{n+1}$ to represent the predictive distribution over the target value $\hat{\mathbf{y}}_{n+1}$.

To learn the parameters of NPs, the loss function is defined by maximizing the evidence lower bound (ELBO) based on the historical FEA simulation data $(\mathbf{x}_1, \mathbf{y}_1), \ldots, (\mathbf{x}_n, \mathbf{y}_n)$ and unobserved target simulation $(\mathbf{x}_{n+1}, \mathbf{y}_{n+1}), \ldots, (\mathbf{x}_{n+T}, \mathbf{y}_{n+T})$, which is shown in equation (4) (Kim et al., 2019).

$$\log p(\mathbf{y}_{n+1:n+T}|\mathbf{x}_{n+1:n+T}, (\mathbf{x}_{1:n}, \mathbf{y}_{1:n})) \geq \mathbb{E}_{q(\mathbf{z}|\mathbf{s}_T)}\left[\sum_{i=1}^{T}\log p(\mathbf{y}_{n+i}|\mathbf{x}_{n+i}, \mathbf{d}_C, \mathbf{z})\right] - D_{KL}(q(\mathbf{z}|\mathbf{s}_T)\|q(\mathbf{z}|\mathbf{s}_C)) \quad (4)$$

In equation (4), $q(\mathbf{z}|\mathbf{s}_T)$ represents the posterior distribution on latent representation $\mathbf{z}$ while $q(\mathbf{z}|\mathbf{s}_C)$ represents the prior on $\mathbf{z}$; $\log p(\mathbf{y}_{n+i}|\mathbf{x}_{n+i}, \mathbf{d}_C, \mathbf{z})$ represents the log probability of true target value on the predictive likelihood; $\mathbf{s}_T$ represents the stochastic representation with both observed and unobserved data points as the input (in the training phase). The detailed derivation of equation (4) is in Appendix II.

## 3. Neural Processes Aided Ordinary Differential Equations

In this section, the limitations of Neural-ODE and NP in modeling FEA simulations and conducting uncertainty quantification will be clarified at first. A novel method, neural processes aided ordinary differential equations (NP-ODE), is then proposed for predictive analytics and uncertainty quantification of FEA simulations. The properties of the proposed NP-ODE will be discussed.

Furthermore, the algorithm of the proposed NP-ODE is developed.

*3.1 Limitations of Neural-ODE and Neural Processes in Modelling FEA Simulations*

As described in section 2.1, Neural-ODE is focusing on using the neural network to parameterize the derivative of the hidden state. In this way, the output of Neural-ODE can be regarded as the integration of inputs over the latent space. While the intuition of Neural-ODE indicates that it is a promising surrogate model of systems governed by differential equations, Neural-ODE is a deterministic model without the uncertainty quantification capability. Because the uncertainties usually exist in complex systems modeled by the FEA method, the deterministic property of Neural-ODE hindered its ability in modeling FEA simulations and capturing the systems' uncertainties.

According to the analysis in section 2.2, NPs have many advantages, such as fitting observed data efficiently, learning predictive distributions instead of a deterministic function, and generating both the mean $\bar{\mathbf{y}}_{n+1}$ and standard deviation $\boldsymbol{\sigma}_{n+1}$ to enable uncertainty quantification of predicted $\hat{\mathbf{y}}_{n+1}$. However, NPs use MLPs as the basic building blocks of its encoder and decoder, which are not the ideal choice, especially in modeling FEA simulations. The reasons can be summarized into three aspects:

(1) The model complexity tends to be significant if we have a large feature vector. For example, suppose MLP stacks $D$ fully connected (FC) layers, and the features' dimensions of each layer are $N$, so that the number of parameters for this MLP is $DN^2$. The number of parameters in the NPs will increase significantly with the higher dimension feature vector;

(2) The dataset containing FEA simulations is usually scarce because the FEA method is usually time-consuming and requires high computational cost. In this case, applying the model with a large number of parameters to model scarce dataset tend to be overfitting;

(3) The FEA is a numerical method to solve PDEs/ODEs that describes a complex system, while the MLP is a purely data-driven method that may lack the physical insights with respect to differential equations.

To this extent, for modeling FEA simulations as well as implement uncertainty quantification, we proposed a novel method, Neural Process Aided Ordinary Differential Equations (NP-ODE), to build an FEA-informed data-driven stochastic surrogate model. Compared with Neural-ODE, our proposed NP-ODE follows the prototype of NPs to capture the uncertainties of predictive results. Compared with NPs, the NP-ODE incorporates Neural-ODE as the decoder, which could not only reduce the number of parameters but also provide physical insights to the surrogates of FEA simulations with respect to differential equations.

## *3.2 General Setup of the NP-ODE to Model FEA Simulations*

In this section, we first formulate the Neural-ODE to model FEA simulations. The basic structure of encoder and decoder in NP-ODE is then illustrated and discussed.

### *3.2.1 Formulate Neural-ODE to Model FEA Simulations*

FEA is a numerical technique to approximate the solutions of partial differential equations or ordinary differential equations (PDEs/ODEs), which are often used to describe a complex system. The surrogate models built upon FEA simulations can reduce the computation cost and resources. Given the introduction in section 2.1, Neural-ODE is a powerful tool in solving systems governed by differential equations so that it has the potential to be an accurate surrogate for FEA methods. The FEA simulations can be formulated as a multivariate regression problem in which the input $\mathbf{x} \in \mathbb{R}^m$ and the output $\mathbf{y} \in \mathbb{R}^p$ variables of FEA are the input and the output of the regression model, respectively. Instead of directly modeling the mapping between $\mathbf{x}$ and $\mathbf{y}$, the Neural-ODE will generate the predicted output by modeling and solving the ODE of the underlying function between $\mathbf{x}$ and $\mathbf{y}$. In this case, suppose the underlying functional relationship between input and output is $\mathbf{y} = \mathbf{f}(\mathbf{x}), \mathbf{x} \in \mathbb{R}^m, \mathbf{y} \in \mathbb{R}^p$, the Neural-ODE will act as a feature extractor, in which the input variables $\mathbf{x} \in \mathbb{R}^m$ are firstly mapped into the feature space $\mathbf{w} \in \mathbb{R}^N$, then the continuous transformation is taken over the feature space, and the output $\mathbf{y}$ in FEA is given from an extra layer with the final features as the input. We select Euler's method as the ODE solver (Griffiths and Higham, 2011). The process of feature transformation is given in equations (5)

$$D_i = D_{i-1} + \Delta D$$
$$\mathbf{g}_{NODE}(D_{i-1}) = \frac{d\hat{\mathbf{f}}_{NODE}(D)}{dD}\bigg|_{D=D_{i-1}}$$
$$\hat{\mathbf{f}}_{NODE}(D_i) \approx \hat{\mathbf{f}}_{NODE}(D_{i-1}) + \Delta D \mathbf{g}_{NODE}(D_{i-1}), \quad i = 1, \ldots, n$$
$$\hat{\mathbf{f}}_{NODE}(D_0) = \mathbf{w} \tag{5}$$

where $\hat{\mathbf{f}}_{NODE}(D)$ represents the approximated discrete feature transformation given by Neural-ODE with a vector as the output, $D$ is equivalent to the depth of Neural Network, $\Delta D$ is the step size. Assume $\hat{\mathbf{f}}_{NODE}(D)$ is continuous and differentiable, $\mathbf{g}_{NODE}(D)$ denotes the first-order derivative of $\hat{\mathbf{f}}_{NODE}(D)$. In this case, $D$ is similar to the number of the intermediate layers in the original NN. As shown in Fig. 2, given the input as $\hat{\mathbf{f}}_{NODE}(D_0)$, the output at model depth $D_i$ is denoted as $\hat{\mathbf{f}}_{NODE}(D_i)$, which can be regarded as a series of discrete transformations. So the value of $D_n$ can be optimized in the training process, which means the Neural-ODE will try to find the most informative feature in the continuous feature transformation.

In summary, the basic steps of building the Neural-ODE as a surrogate model of FEA include: a) formulate FEA simulations as a regression with input $\mathbf{x} \in \mathbb{R}^m$ and output $\mathbf{y} \in \mathbb{R}^p$; b) map the input variables $\mathbf{x} \in \mathbb{R}^m$ into feature space $\mathbf{w}_{D_0} \in \mathbb{R}^N$; c) extract the most informative features from the continuous transformation modeled by Neural-ODE; d) map the extracted features into the output space $\mathbf{y} \in \mathbb{R}^p$.

*3.2.2 Structure of Encoder in NP-ODE*

The basic structure of NP-ODE follows the prototype of the NPs, which mainly consists of an encoder and a decoder. The role of the encoder is to extract the deterministic and stochastic representations of observed data points. Its structure is given in Fig. 4, in which the main components are the deterministic encoder, the stochastic encoder, and the attention module.

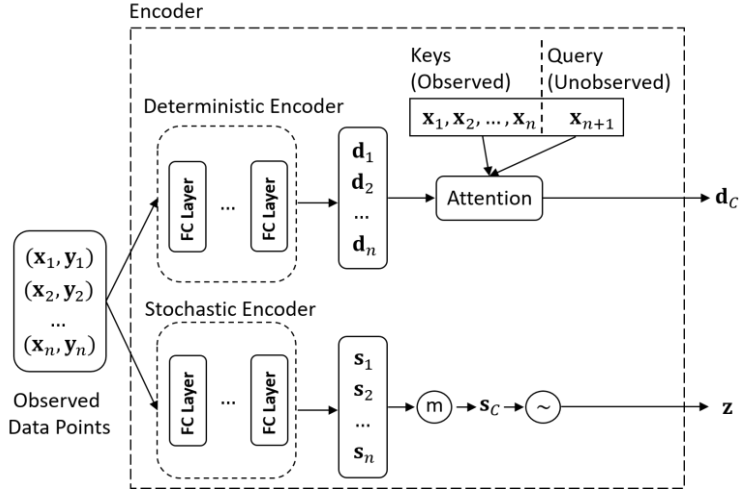

Figure 4. Structure of Encoder in the NP-ODE

The deterministic and stochastic encoders share a similar structure, which are consists of MLP (stacked FC layers). The expressions of deterministic and stochastic encoders are given in equations (6).

$$\begin{aligned}\mathbf{d_i} &= <\mathbf{W}_{DE}, (\mathbf{x}_i, \mathbf{y}_i)> + \mathbf{b}_{DE}\\ \mathbf{s_i} &= <\mathbf{W}_{SE}, (\mathbf{x}_i, \mathbf{y}_i)> + \mathbf{b}_{SE}\end{aligned} \quad (6)$$

In equations (6), $\mathbf{W}_{DE}, \mathbf{b}_{DE}$ denote the weight matrix and bias in the FC layer of the deterministic encoder, $\mathbf{W}_{SE}, \mathbf{b}_{SE}$ denote the weight matrix and bias in the FC layer of the stochastic encoder, $\mathbf{d_i}, \mathbf{s_i}$ are the deterministic and stochastic representations of observed data point $(\mathbf{x}_i, \mathbf{y}_i)$.

The role of the attention module here is to take the weighted aggregation of the deterministic representations $(\mathbf{d}_1, \dots, \mathbf{d}_n)$ based on the similarity of observed inputs $(\mathbf{x}_1, \dots, \mathbf{x}_n)$ and the queried (unobserved) input $\mathbf{x}_{n+1}$. The multi-head attention module is applied in the NP-ODE (Vaswani et al. 2017). It is worth noting that the output of the attention module has the permutation invariant property so that the order of deterministic representations $(\mathbf{d}_1, \dots, \mathbf{d}_n)$ and observed inputs $(\mathbf{x}_1, \dots, \mathbf{x}_n)$ does not influence the output $\mathbf{d}_c$. Similarly, the $\mathbf{s}_c$ is generated by the mean aggregation of stochastic representations $(\mathbf{s}_1, \dots, \mathbf{s}_n)$, which also has the permutation invariant property.

*3.2.3 Structure of Decoder in NP-ODE*

The NP-ODE is built to incorporate the Neural-ODE as the decoder into NPs, which not only strengthened the ability in modeling FEA simulations but also gained the property to capture system uncertainties.

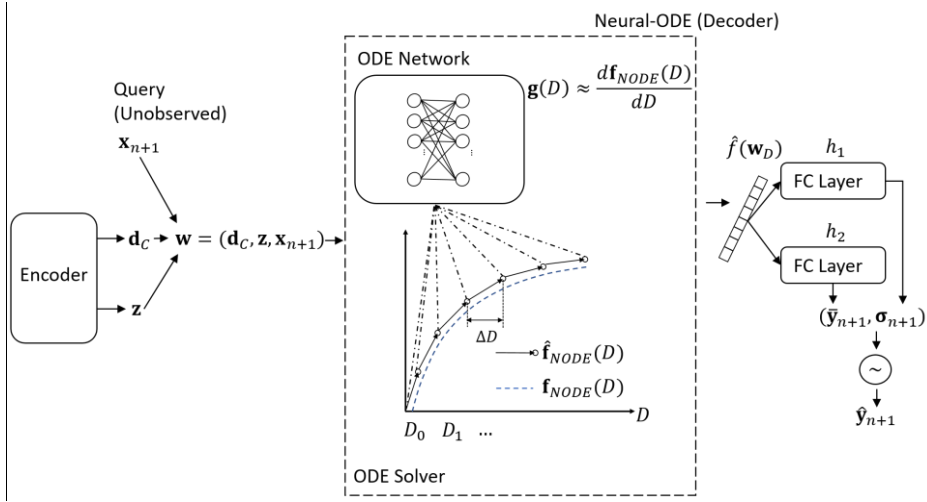

Figure 5. Structure of Decoder in the NP-ODE

As Fig. 5 shows that in the Neural-ODE, the ODE network along with ODE solver (e.g. Euler method) could approximate the continuous transformation $\mathbf{f}_{NODE}(D)$ of the input representations, and a smaller step size $\Delta D$ could give a better approximation accuracy of FEA simulations. In contrast, the MLP decoder in the NPs stacks multiple FC layers which can be regarded as the discretization of the continuous transformations. Intuitively, increasing the number of intermediate layers in the MLP can give a better approximation, however, this will increase the number of parameters significantly. To this extent, similar to original Neural Networks, the Neural-ODE also has the property equivalent to the model depth (number of intermediate layers), which is determined by the step size $\Delta D$. As long as the step size $\Delta D$ is small enough and the predicted derivative at each step given by the ODE network is accurate enough, the Neural-ODE is equivalent to an infinite-layer Neural Network. The advantage of Neural-ODE is that it uses the same ODE network to generate the derivative of the underlying function at each step, which reduces the number of parameters apparently compared with MLP.

In general, the expressions of the decoder in NP-ODE are summarized as equations (7).

$$\hat{\mathbf{y}}_{n+1} \sim \mathcal{N}(\bar{\mathbf{y}}_{n+1}, \boldsymbol{\sigma}_{n+1})$$
$$\bar{\mathbf{y}}_{n+1} = h_1\left(\hat{\mathbf{f}}_{NODE}(D_n)\right)$$
$$\boldsymbol{\sigma}_{n+1} = h_2\left(\hat{\mathbf{f}}_{NODE}(D_n)\right)$$
$$\hat{\mathbf{f}}_{NODE}(D_i) \approx \hat{\mathbf{f}}_{NODE}(D_{i-1}) + \Delta D \mathbf{g}_{NODE}(D_{i-1})$$
$$\hat{\mathbf{f}}_{NODE}(D_0) = \mathbf{w} = (\mathbf{d}_C, \mathbf{z}, \mathbf{x}_{n+1})$$

$$\mathbf{g}(D_{i-1}) = \frac{d\hat{\mathbf{f}}_{NODE}(D)}{dD}\bigg|_{D=D_{i-1}}$$
$$D_i = D_{i-1} + \Delta D, i = 1, \dots, n \tag{7}$$

In equations (7), $h_1(.)$ and $h_2(.)$ represent two FC layers used to map the output features from Neural-ODE to predicted mean $\bar{\mathbf{y}}_{n+1}$ and standard deviation $\boldsymbol{\sigma}_{n+1}$, $\hat{\mathbf{f}}_{NODE}(D_i)$ represents the dynamic feature transformations modeled by Neural-ODE, $\mathbf{d}_C$ and $\mathbf{z}$ remain the same as NPs which are the deterministic and stochastic representations of observed data respectively. As the equations (7) show that the NP-ODE is designed to model systems governed by the differential equations. As a proof-of-concept, a simulation study is conducted in section 4 to validate the strength of the NP-ODE in modeling data points governed by differential equations compared with the original NPs.

Combine the introductions of encoder and decoder in sections 3.2.2 and 3.2.3, the proposed NP-ODE is ready to use. Before demonstrating the performance of NP-ODE, uncertainty quantification and its properties will be further analyzed.

## *3.3 Uncertainty Quantification*

Uncertainty quantification (UQ) conducted by our proposed NP-ODE is to quantitatively measure the uncertainties of predicted results given new FEA inputs. We take the uncertainties in the FEA for Tribocorrosion (introduced in section 5.1) as an example. There are six parameters representing material properties used as the inputs of FEA. The FEA method for Tribocorrosion is built to deterministically simulate the corrosion rate under various combinations of parameters. However, in reality, these parameters are not deterministic in tribocorrosion FEA simulation because: (1) the corrosion process changes the properties of the material; (2) given the same set of parameters, the corrosion rate (FEA output) may have slight changes in different runs of real experiments. Since the FEA method can only consider a limited number of critical parameters, other environmental parameters may inevitably influence the corrosion rate. The idea of uncertainty quantification in our proposed NP-ODE lies in capturing the variations of FEA simulations under various combinations of parameters. So that the output of NP-ODE is a distribution of the predicted result instead of a specific value, which represents how likely is the predicted output. We select the confidence interval to numerically measure the uncertainties of each output, which means considering the systems' uncertainties, the corrosion rate is likely to locate within the predicted interval.

## *3.4 Properties of NP-ODE*

The properties of our proposed NP-ODE can be summarized into three folds.

Firstly, compared with the MLP decoder, incorporating Neural-ODE as the decoder can reduce the number of parameters because it eliminates the multi-layer structure and repetitively apply the same ODE network to generate the derivatives at each step. The number of parameters in the $D$-layer MLP is $DN^2$, in which $N$ is the feature dimension, $D$ is equivalent to the number of steps in discrete transformations. The parameters in the corresponding Neural-ODE is not influenced by the number of steps, because the same ODE network will be applied at each step to estimate the derivatives. If the ODE network consists of FC layers, the number of parameters in the Neural-ODE could be expressed as $N^2$. More importantly, other types of layers (e.g. convolutional (Conv) layer) can also be used as the building block in the ODE network. Compared with the FC layer, the number of parameters in the Conv layer will not be determined by the feature dimension $N$, which enables the Neural-ODE to further reduce parameters. In our experiments, the Conv layer is selected as the basic building block in the ODE network. A detailed comparison between the number of parameters will be introduced in Section 5.3.

Secondly, the NP-ODE and the FEA method share a similar idea in emulating complex systems, which is modeling and solving the PDEs/ODEs that represent the complex system, instead of directly modeling the functional relationship between input and output by pure data-driven models.

Thirdly, the proposed NP-ODE captures the output uncertainties and generates the distribution over outputs for FEA simulations. Given the output distribution, we can further conduct uncertainty quantification on outputs by generating confidence intervals at each data point.

## *3.5 Pseudo-code of the Algorithm for NP-ODE*

The pseudo-code of applying NP-ODE to model FEA simulations and implement uncertainty quantification is summarized in Algorithm 1, in which $\mathcal{D}_C$ represents the dataset containing inputs and outputs of historical FEA simulations, $\mathcal{D}_T$ represents the dataset containing target data points (unobserved FEA simulation), $x_T$ represents the dataset containing target FEA inputs.

---

**Algorithm 1**: NP-ODE in modeling FEA simulations and conducting uncertainty quantification

---

**Inputs:**
1: $\mathbf{x} \in \mathbb{R}^m, \mathbf{y} \in \mathbb{R}^p$    ▷ data from FEA simulations
2: $\mathcal{D}_{train} = (\mathbf{x}_1, \mathbf{y}_1), \dots, (\mathbf{x}_n, \mathbf{y}_n)$    ▷ training data
   $\mathcal{D}_{test} = (\mathbf{x}_{n+1}, \mathbf{y}_{n+1}), \dots, (\mathbf{x}_{n+T}, \mathbf{y}_{n+T})$    ▷ testing data

**Training:**
3: **while** $i \leq$ number of iterations **do**
4:     randomly select $\mathcal{D}_C \subseteq \mathcal{D}_{train}, \mathcal{D}_T = \mathcal{D}_{train} - \mathcal{D}_C$
5:     prior $q(\mathbf{z}|\mathbf{s}_C) \leftarrow$ feed $\mathcal{D}_C$ into stochastic encoder
6:     posterior $q(\mathbf{z}|\mathbf{s}_T) \leftarrow$ feed $\mathcal{D}_{train}$ into stochastic encoder
7:     deterministic representation $\mathbf{d}_C \leftarrow$ feed $(\mathcal{D}_C, x_T)$ into the deterministic encoder
8:     **for** $(\mathbf{x}_t, \mathbf{y}_t)$ in $\mathcal{D}_T$ **do**
9:         sample $\mathbf{z}$ from $q(\mathbf{z}|\mathbf{s}_T)$
10:        $\bar{\mathbf{y}}_t, \boldsymbol{\sigma}_t \leftarrow$ feed $(\mathbf{z}, \mathbf{d}_C, \mathbf{x}_t)$ into decoder
11:        $\hat{\mathbf{y}}_t \sim \mathcal{N}(\bar{\mathbf{y}}_t, \boldsymbol{\sigma}_t)$
12:        calculate likelihood $p(\mathbf{y}_t|\mathbf{x}_t, \mathbf{d}_C, \mathbf{z})$
13:    **end for**
14:    $\mathcal{L} = \mathbb{E}_{q(\mathbf{z}|\mathbf{s}_T)}[\sum_{t=1}^{T} \log p(\mathbf{y}_t|\mathbf{x}_t, \mathbf{d}_C, \mathbf{z})] - D_{KL}(q(\mathbf{z}|\mathbf{s}_T) \| q(\mathbf{z}|\mathbf{s}_C))$
15:    update parameters to maximize $\mathcal{L}$
16:    $i = i + 1$
17: **end while**

**Testing:**
18: $\mathcal{D}_C = \mathcal{D}_{train}, \mathcal{D}_T = \mathcal{D}_{test}$
19: prior $q(\mathbf{z}|\mathbf{s}_C) \leftarrow$ feed $\mathcal{D}_C$ into stochastic encoder
20: deterministic representation $\mathbf{d}_C \leftarrow$ feed $(\mathcal{D}_C, x_T)$ into the deterministic encoder
21: **for** $(\mathbf{x}_t, \mathbf{y}_t)$ in $\mathcal{D}_T$ **do**
22:    sample $\mathbf{z}$ from $q(\mathbf{z}|\mathbf{s}_C)$
23:    $\bar{\mathbf{y}}_t, \sigma_t \leftarrow$ feed $(\mathbf{z}, \mathbf{d}_C, \mathbf{x}_t)$ into decoder
24:    $\hat{\mathbf{y}}_t \sim \mathcal{N}(\bar{\mathbf{y}}_t, \sigma_t)$
25:    generate confidence interval using predicted distribution

---

In this section, we introduced the general set up of our proposed method NP-ODE along with its expressions, analyzed its properties, and summarized the pseudo-code to illustrate its training and testing procedures. In summary, the advantages of the NP-ODE include a) compare with NPs, it improves the parameter efficiency to reduce the number of parameters in the decoder; b) compare with Neural-ODE, it generates the distribution over the predicted output to enable uncertainty quantification; c) it shares the similar differential equations structure with FEA method in emulating complex systems so that it is more promising to be a surrogate model of FEA simulations.

## 4. Simulation study

To validate the effectiveness of our proposed NP-ODE in solving ODEs, a simulation study is conducted by restoring dynamic functions using sampled data from a given ODE and conducting uncertainty quantification. Furthermore, to demonstrate the robustness of the NP-ODE, the experiment is repetitively conducted on sampled data with different levels of Gaussian random noise. In this section, the simulation setup is first introduced to illustrate how the sampled data is collected. The evaluation metrics are then discussed. Finally, the benchmark method is introduced, and performances comparison between the NP-ODE and the benchmark method are demonstrated and discussed.

### *4.1 Simulation setup*

The two-dimensional spirals are selected to generate simulated data points, which can be modeled by a linear ODE as shown in equation (8).

$$\frac{d\mathbf{y}}{dx} = \begin{bmatrix} -0.1 & -1 \\ 1 & -0.1 \end{bmatrix} \mathbf{y} \tag{8}$$

In which, $x \in \mathbb{R}$ is the input and can be regarded as the order of data points, $\mathbf{y} \in \mathbb{R}^2$ is the two-dimensional output and can be regarded as the position of data points.

As we discussed in section 2.1, data points following equation (8) can be generated by equation (9), in which $\boldsymbol{\epsilon} \in \mathbb{R}^2$ is the Gaussian random noise controlling the variations of generated data points.

$$\mathbf{y}_N = 4 \times \text{ODESolve}\left(\mathbf{y}_0, \frac{d\mathbf{y}}{dx}, x_0, x_N\right) + \boldsymbol{\epsilon} \tag{9}$$

Our proposed NP-ODE is robust in modeling data points with various levels of variations. To demonstrate the robustness, the mean of $\boldsymbol{\epsilon}$ is set as $0$, and the standard deviation of $\boldsymbol{\epsilon}$ is set as $0.01, 0.02, 0.1$ respectively. For each noise level, 200 data points are generated for training and validating the method. The generated data points are visualized in Fig. 6, in which the first subplot from the left shows the spiral without noise, the scatters in the rest three subplots are data points sampled from this spiral with various noise levels. We can find out that when the standard deviation of noise equals 0.1, the pattern of the sampled data points is hard to visualize.

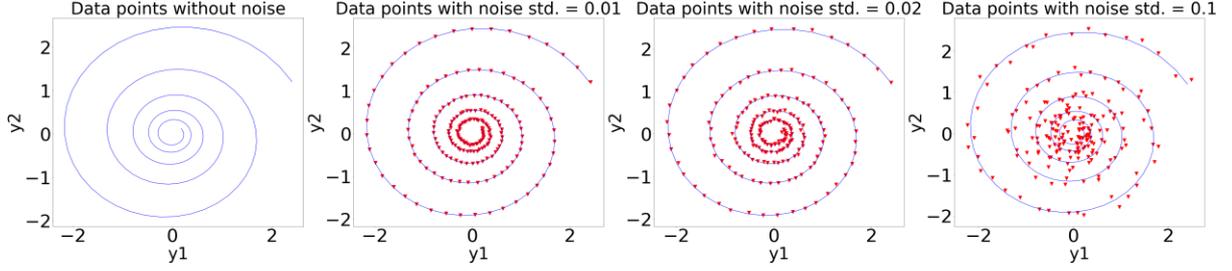

Figure 6. Visualization of generated data points

*4.2 Evaluation Metrics*

To demonstrate the model's ability in identifying the underlying patterns of noisy data points sampled from an ODE, the rooted mean square error (RMSE) is selected to evaluate the predictive accuracy and the confidence interval (CI) within one standard deviation is selected to conduct uncertainty quantification.

The expression of RMSE is given in equation (10), in which $\mathbf{y}_i$ is the real position of the $i_{th}$ data point, $\bar{\mathbf{y}}_i$ is the predicted position of the $i_{th}$ data point, and $\|.\|_2^2$ is the square of the $l_2$ norm. A smaller value of RMSE on test points indicates the model can predict more accurately,

$$\text{RMSE} = \sqrt{\frac{1}{2N}\sum_{i=1}^{N}\|\mathbf{y}_i - \bar{\mathbf{y}}_i\|_2^2} \quad (10)$$

The CI within one standard deviation is selected for uncertainty quantification, which is denoted as $(\bar{\mathbf{y}}_i - \hat{\boldsymbol{\sigma}}_i, \bar{\mathbf{y}}_i + \hat{\boldsymbol{\sigma}}_i)$, and $\hat{\boldsymbol{\sigma}}_i \in \mathbb{R}^2$ is the predicted standard deviation in two dimensions. To show the robustness of the proposed NP-ODE, it should recover the underlying patterns (spiral without noise) from noisy data points. In this case, if the generated CI can cover the spiral curve, it indicates the corresponding method is more robust to the noise.

*4.3 Comparison with Benchmark Methods*

To show the robustness of the proposed NP-ODE in modeling data points with different noise levels, we compare its performance on the simulated data points to the original NPs. The detailed introduction of NPs is given in section 2.2. In this section, we introduce the details of the experiment and discuss the results from the perspectives of mean prediction performance and uncertainty quantification, respectively.

*4.3.1 Experiment Design*

As shown in Fig. 6, there are 3 noise levels of the simulated data with the standard deviation of noise equalling 0.01, 0.02, 0.1, respectively. For each noise level, 200 data points are generated, in which 150 data points are randomly selected to train the model, and the rest 50 data points are selected to test the model. To keep a fair comparison, the details of experiments using NP-ODE and NPs keep the same. During the training phase, part of the training data are randomly selected as the context (observed) data points with both the input $x_i$ and response $\mathbf{y}_i$ feeding into the model, and the rest training data are target (unobserved) data points with only input $x_i$ feeding into the model. During the testing phase, all the training data will serve as the context data points, and the testing data will be target data points. The experiment will be conducted on three noise levels of the simulated data separately.

*4.3.2 Mean Prediction Comparison*

The mean prediction comparison on test data points between the proposed NP-ODE and the NPs are shown in Table 1. We can conclude that under different noise levels, the proposed NP-ODE can consistently generate more accurate predictions compared with the original NPs. Furthermore, with the standard deviation of noise increases, the prediction improvement is more significant.

Table 1 Results Comparison (RMSE)

| Standard deviation of noise | 0.01 | 0.02 | 0.1 |
|---|---|---|---|
| NPs | 0.0146 | 0.0268 | 0.2778 |
| NP-ODE | **0.0136** | **0.0256** | **0.1171** |

*4.3.3 Uncertainty Quantification Comparison*

To compare the robustness of the proposed NP-ODE with the original NPs, in Fig. 7, we plot the generated curves on both training and testing data points along with the CI within one standard deviation. In Fig. 7, the red rectangles denote testing data points and the green stars denote training data points; the blue curves are spiral without noise, which is the underlying pattern of noisy data points; the red curves are generated by the NP-ODE or NPs on both training and testing data points; the cyan shadow represents the CI within one standard deviation in the dimension of $y_1$, and yellow shadow represents the CI within one standard deviation in the dimension of $y_2$. The left and right columns of Fig. 7 contain the results from the NP-ODE and the NPs, respectively. The first row indicates the training and testing data points (with noise) and the underlying spiral without noise, which are kept the same in training the NP-ODE and NPs. The middle two rows compare the NP-ODE and the NPs from perspectives of UQ in each dimension $y_1$ and $y_2$, respectively. The last row can be regarded as the combination of the middle two rows, which shows that compared with the NPs, the proposed NP-ODE successfully captures the patterns of noisy data points, and the generated CI covers the underlying spiral curve. Note that Fig. 7 only shows the modeling results on noisy data points with the noise standard deviation as 0.1. The results on other noise levels are listed in Appendix III.

Fig. 7 shows that the generated trajectory from the proposed NP-ODE is consistent with the actual spiral curve, which demonstrates the NP-ODE successfully reveals the underlying data properties. Via Table 1 and Fig. 7, we can conclude that the proposed NP-ODE shows its strength in modeling data points governed by differential equations and demonstrates the robustness in recovering underlying patterns with noisy data points.

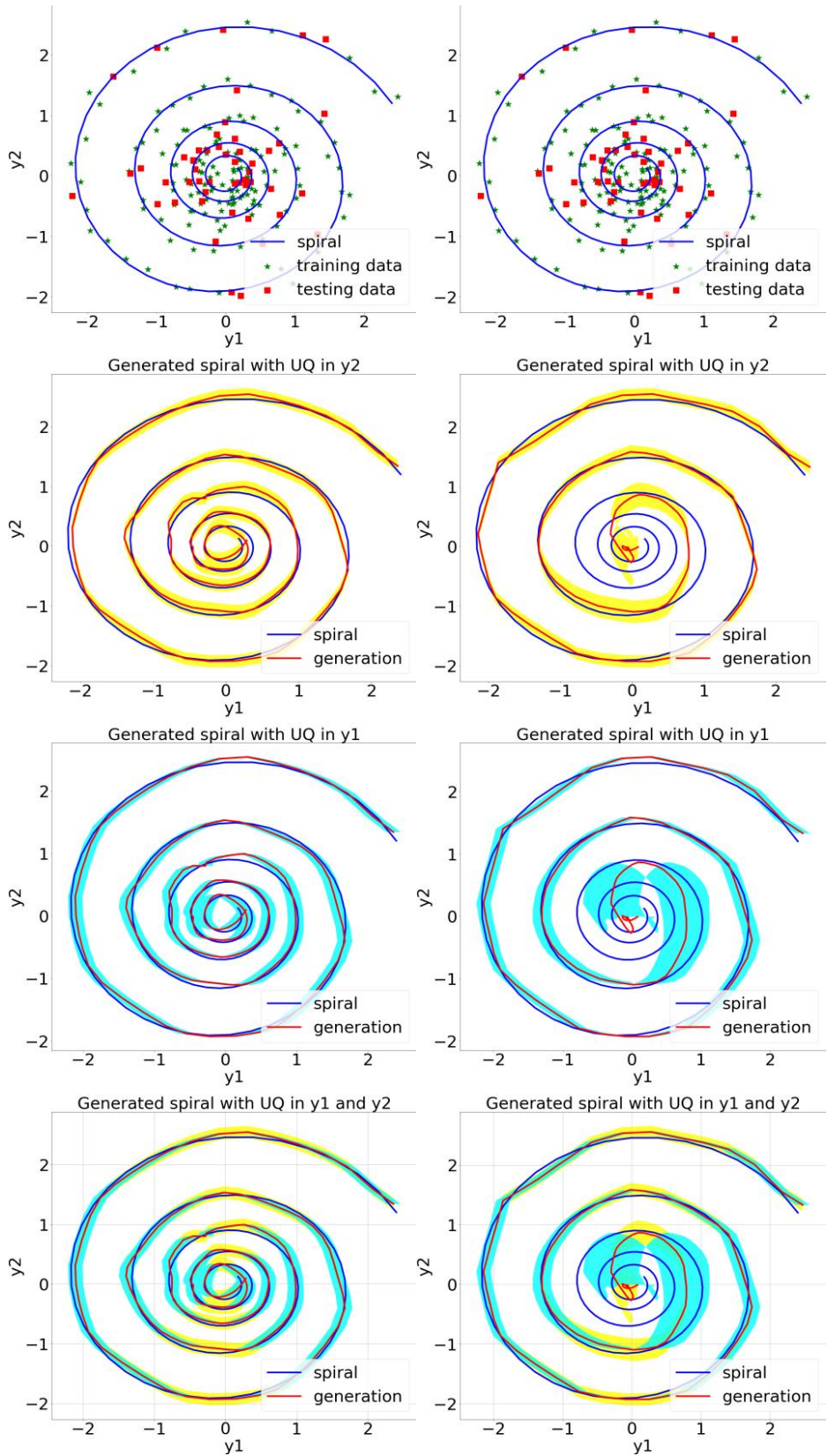

Figure 7. UQ Comparison with noise $\sigma = 0.1$; **Left:** result from NP-ODE; **Right**: result from NPs

# 5. Case study

To validate and demonstrate the strength and effectiveness of the proposed NP-ODE, a case study is conducted to build a surrogate model of FEA simulations and do uncertainty quantification. In this section, at first, we introduced FEA simulations on material corrosion analysis during new materials design, and preprocess the data for surrogate models. Furthermore, the implementation details of NP-ODE on the specific dataset are discussed including the model structure and parameters analysis. Additionally, evaluation metrics are selected to measure model performance. Finally, benchmark models are introduced and the comparisons among our proposed model and benchmark models are discussed.

## *5.1 Introduction to FEA for Tribocorrosion*

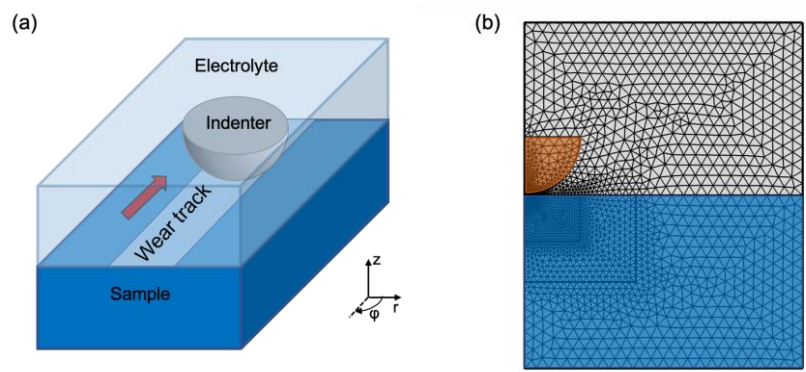

Figure 8. (a) Schematic of FEA for Tribocorrosion Test; (b) Meshing Setup of Tribocorrosion Test

Tribocorrosion is a material degradation process involving both mechanical wear and corrosion of the material, which jeopardizes materials' long-term sustainability and structural integrity. Tribocorrosion analysis is very important for design and manufacturing systems. The synergetic effects of mechanical damage and corrosion could cause more severe material degradation than the sum of pure wear and corrosion. To investigate the effects of materials' mechanical and electrochemical properties on their tribocorrosion behavior, an FEA model with the capability of simulating both the dynamic process of wear and the time-dependent evolution of corroding surface during tribocorrosion is developed and verified by an experimental study of two aluminum alloys (with 5 wt% Mn and 20 wt% Mn respectively). The scheme of FEA and the meshed geometry are shown in Fig. 8. The model first simulates a scratching wear process and produce results including the wear volume loss, and surface and subsurface stress and strain caused by the process. A phenomenological model reflecting the change in anodic potential caused by plastic strain is then used to incorporate the wear-corrosion synergy. With the impact on the electrochemical state of the system caused by mechanical deformation considered, the corrosion process is simulated.

## *5.2 Dataset Introduction and Preprocessing*

To investigate the influence of material properties' on tribocorrosion rate, six individual parameters were taken into consideration, including mechanical (Young's modulus and yield strength) and corrosion (anodic and cathodic Tafel slope and exchange current density) parameters. The final output of the model is the material loss rate caused by tribocorrosion. In this specific case, the material loss caused by corrosion and wear-corrosion synergy is of interest, which can also be expressed as total material loss due to tribocorrosion minus that due to pure wear.

The effects of mechanical, anodic corrosion and cathodic corrosion parameters were investigated separately. For the variation of mechanical parameters, Young's modulus was swept from 55 MPa to 95 MPa with a 5 MPa step size, the yield strength from 1.0 MPa to 5.0 MPa with a 0.5 MPa step size, while the corrosion parameters were kept constant. For the variation of corrosion parameters, the cathodic Tafel slope was varied from -280 to -210 mV/decade, the cathodic exchange current density from $2.0 \times 10^{-8}$ to $2.0 \times 10^{-7}$ A/cm$^2$, the anode Tafel slope from 250 to 290 mV/decade, and the anodic exchange current density from $1.0 \times 10^{-13}$ to $5.0 \times 10^{-13}$ A/cm$^2$. Each combination of mechanical parameters and corrosion parameters will render a different output of material loss rate.

Given the dataset generated from FEA, the data preprocessing is further applied to make it ready for building surrogate models. In the dataset, there are 106 simulations and each simulation contains six input variables representing the mechanical and corrosion properties and one output variable representing the corrosion rate. Considering the input and output variables have different physical meanings and significant various scales, the normalization is applied to the raw data to transform all the variables into $[-2,2]$, the expression of normalization is shown in equation (11).

$$x_{\text{norm}} = \left(\frac{x - x_{\min}}{x_{\max} - x_{\min}} \times 4\right) - 2 \tag{11}$$

In equation (11), $x$ represents a single variable, $x_{\max}$ and $x_{\min}$ are the maximum and minimum value of $x$ respectively. After normalization, 20 simulations are randomly selected and fixed as the testing data and the rest 86 simulations are training data.

## 5.3 Parameters Analysis and Selection

Given the data from FEA simulations, in this section, the implementation details of the proposed NP-ODE will be first introduced, which includes the selection of ODE network in the decoder of NP-ODE; the number of parameters will be further analyzed and compared with original NPs given the specific model structure; the selection of parameters in the Euler's method for solving NP-ODE is also discussed.

Compare with the original NPs, the NP-ODE uses fewer parameters because of incorporating the Neural-ODE as the decoder. In our experiment, the convolutional layer (Conv layer) is selected as the basic building block of Neural-ODE to capture the derivative of continuous feature transformation. The reasons can be summarized into three aspects. Firstly, the original NPs use the MLP as the decoder, in which the number of parameters is determined by the dimensions of input and output feature vectors. In contrast, the number of parameters in the Conv layer is influenced by the size and the number of convolutional filters which is not determined by the input and output. To this extent, especially for those high dimensional features, the choice of the Conv layer could reduce the number of parameters. Secondly, the Neural-ODE could use limited layers to model the derivative and further mimic the continuous transformation, while the MLP in NPs is to discretize the continuous transformation by the multi-layer structure. In this case, the number of parameters in NPs will be determined by the model depth while the depth in the Neural-ODE will not influence parameters. Finally, when comparing the single Conv layer and FC layer (the building block of MLP), the choice of the Conv layer may be hindered because it only captures local features limited by the size of the convolutional filter. However, since the Neural-ODE can be regarded as the infinite-layer neural network, the global feature can be captured hierarchically, given the infinite number of Conv layer.

Table 2. Comparison of Decoder Parameters Between NPs and NP-ODE

| Model | Decoder Layer | Weight Matrix | # Parameters |
|---|---|---|---|
| NPs | FC layer × 3 | (384,384) | 147456 |
|  |  | (384,384) | 147456 |
|  |  | (384,384) | 147456 |
| NP-ODE | Conv layer 1 | (1,128,3,1) | 384 |
|  | Conv layer 2 | (129,128,3,1) | 49536 |
|  | Conv layer 3 | (129,128,3,1) | 49536 |

For example, in the decoder of NP-ODE, we use three Conv layers to build the ODE network to estimate the derivatives of feature transformation. Combine the ODE network with the ODE solver, the Neural ODE can mimic the continuous feature transformation, while in the decoder of NPs, we use three FC layers to discretize the continuous feature transformation. Suppose the input feature of the decoder is of shape (1,384,1), the detailed model structure, shape of the weight matrix of each layer, and the number of parameters of each layer are summarized in Table 2. From the comparison, we can find that the number of parameters in the decoder of NPs is $147456 \times 3 = 442368$ while the number of parameters in the decoder of NP-ODE is $384 + 49536 \times 2 = 99456$. Furthermore, the number of parameters in the decoder of NPs will further increase with the increase of model depth, while the model depth will not influence the number of parameters in the decoder of NP-ODE. Given the dataset generated from FEA simulations are often scarce because of the high cost of data collection, fewer parameters in the proposed NP-ODE reduce the possibility of overfitting.

Euler's method is selected as the ODE solver in our experiment as shown in equation (7). The $D_0, D_N, \Delta D$ are important parameters influencing the performance of Euler's method, which denote the input model depth, output model depth, and step size, respectively. Theoretically, the difference between the input and output depth $(D_N - D_0)$ along with the step size $\Delta D$ determine the result accuracy of Euler's method. Given a specific value of $(D_N - D_0)$, a smaller step size will improve the result accuracy and increase the training time. Considering such a trade-off between accuracy and efficiency, we select the $(D_N - D_0)$ as 1 and $\Delta D$ as 0.05, which are proved to be a good combination given the model's performances in prediction accuracy (sections 5.5.3, 5.5.4) and computational cost (section 5.5.5).

## 5.4 Evaluation Metrics

To evaluate the performance of our proposed model and compare it with benchmark methods, two evaluation metrics are selected and applied to the testing data, which are mean absolute percentage error (MAPE) for predictive accuracy, and 95% confidence interval (CI) for uncertainty quantification.

The expression of MAPE is given as equation (12). The MAPE is a relative error evaluating the relative difference between the predicted error and real value, which can be regarded as eliminating the differences among data scales and treating each data point equally.

$$\text{MAPE} = \frac{1}{N} \sum_{i=1}^{N} \left| \frac{y_i - \bar{y}_i}{y_i} \right| \qquad (12)$$

In equation (12), $N$ represents the number of testing data points, $\bar{y}_i$ is the predicted mean value of $i_{th}$ data point.

The output of NP-ODE is the distribution of predicted value which consists of predicted mean $\bar{y}$ and predicted standard deviation $\hat{\sigma}$. The confidence interval can be generated for uncertainty quantification. We select the 95% confidence interval, which is calculated by $(\bar{y} - 1.96\hat{\sigma}, \bar{y} + 1.96\hat{\sigma})$ for each testing data point. There are three criteria to evaluate the quality of confidence interval: a) the real value of testing data point lies in the confidence interval of predicted value; b) the real value is close to the center of confidence interval (predicted mean $\bar{y}$); c) a smaller $\hat{\sigma}$ ensuring the coverage of the real value of FEA simulations.

## 5.5 Comparison with Benchmark Methods

To validate the performance of our proposed NP-ODE, we apply the proposed NP-ODE and other benchmark methods to build surrogates of the FEA simulations and compare their performance. The original Neural Processes, Gaussian Process with Matern kernel, and Gaussian Process with Polynomial kernel are selected as benchmark methods. In this section, we will firstly give a brief introduction of these benchmark methods, illustrate the design of the experiment, and compare models' performances on the FEA simulations.

*5.5.1 Introduction of Benchmark Methods*

The first benchmark method is the original NPs and it is selected to show the advantages of incorporating Neural-ODE as the decoder. The details of the NPs are introduced in Section 2.2. The rest two benchmark methods belong to the family of the Gaussian Process. In the Gaussian Process, instead of finding a deterministic function, it will derive the probability distribution over all possible functions that fit the data. The basic steps are a) to specify a prior distribution on the function space; b) according to the Bayesian rule, to calculate the posterior distribution using training data; c) to use posterior distribution to make inference on testing data. The choice of covariance function is a significant factor influencing the performance of the Gaussian Process. We selected Gaussian Processes with Matern and Polynomial kernels as the second and third benchmark methods. These two kernel functions are commonly used to capture non-linear relationships.

*5.5.2 Experiment Design*

In the experiment, each FEA simulation can be formulated as a data point $(\mathbf{x}, y), \mathbf{x} \in \mathbb{R}^6, y \in \mathbb{R}$. Among all the 106 simulations, 20 simulations are randomly selected and fixed as the testing data, and the rest 86 simulations are training data. To test the model's ability to explore limited data points, we repeat the experiment, change the number of training points available to models, and test on the same testing data to check the robustness of models' performances. To ensure a fair comparison, the number of training samples is iteratively increased, which is 30 samples out of 86 training data are randomly selected at first, and then the sample size is increased to 50, 60, 70, and 80 by successively adding new data points from training data randomly. The comparison of results among all the methods is introduced next.

*5.5.3 Mean Predictions Comparison*

The outputs of our proposed NP-ODE and benchmark methods are all distributions over predicted value. To numerically evaluate the predicted results, we calculate the MAPE between the predicted mean $\bar{y}$ and real value $y$ on testing data. The results are summarized in Table 3.

Table 3. Results Comparison (MAPE %)

| # Train | NP-ODE | NPs | GP (Matern) | GP (Poly) |
|---|---|---|---|---|
| 30 | **5.5229** | 9.2862 | 11.8461 | 19.3152 |
| 50 | **2.7286** | 4.9690 | 7.1928 | 10.3434 |
| 60 | **2.5886** | 4.8660 | 6.6770 | 14.1264 |
| 70 | **2.2121** | 3.2018 | 5.1234 | 2.3593 |
| 80 | **2.1235** | 3.2364 | 4.4351 | 2.4418 |

In Table 3, the first column represents the number of training samples available to models, and each row is the testing MAPE of all models. By comparing the results from each row, we can conclude that given the same training samples, our proposed NP-ODE outperforms benchmark methods consistently. By comparing the results from each column, we can conclude that with more training samples, all the methods tend to have a better performance. Furthermore, the performance of NP-ODE has the smallest variation and the best accuracy with the different number of training samples, which means our proposed model can explore the features in training samples better and has a more robust performance with the different number of training samples.

*5.5.4 Uncertainty Quantification Comparison*

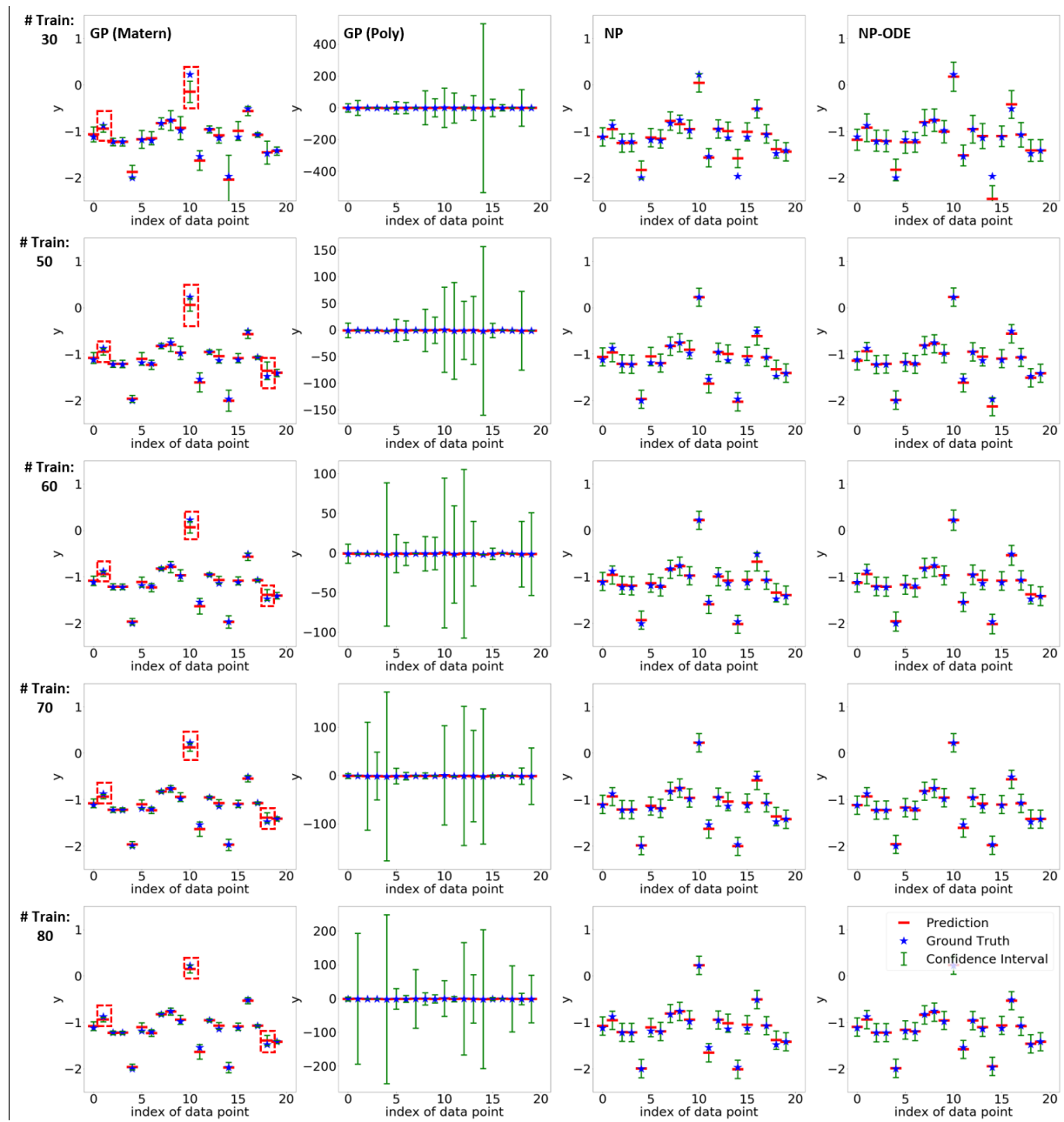

Figure 9. Visualization of Uncertainty Quantification

Given the predicted mean and standard deviation on each testing data point, the 95% confidence interval is generated for uncertainty quantification. The visualization of uncertainty quantification is shown in Fig. 9, in which the blue star represents the ground-truth value, the red line represents the predicted mean, and the green-capped line represents the 95% confidence interval ($\pm 1.96\hat{\sigma}$). From Fig. 9, we can find that (a) with more training data, the predicted mean value (red line) tends to get closer to the real value (blue star); (b) the GP with Matern kernel tends to underestimate the standard deviation, and some of the real value lies out of the boundary of 95% confidence interval (highlighted in red dashed boxes); (c) the GP with Polynomial kernel tends to overestimate the standard deviation, which makes the confidence interval have no sense in applications; (d) the NPs and NP-ODE generally generate a reasonable standard deviation and predicted mean at each data point, which makes the real value close to the predicted mean and the confidence interval gives a reasonable estimation of uncertainty.

From the comparison of uncertainty quantification among these four methods, we can conclude that the NPs and the NP-ODE can give a more reasonable estimation of both the mean and standard deviation than the Gaussian Process. The NP-ODE keeps the real value close to the center of the confidence interval and generate a reasonable estimation of uncertainty. Furthermore, the NP-ODE gives the most accurate prediction on testing data points and when feeding 80 training samples into the model, the predicted 95% confidence interval has the best coverage on testing data points.

*5.5.5 Computational Cost*

The training time of the proposed NP-ODE is mainly determined by the number of iterations required for training. The iteration here denotes one complete forward and backward propagation to update the model parameters. The experiments are conducted on a single NVIDIA TITAN V GPU, and it takes around 0.1 seconds to run a single iteration in training NP-ODE. The training process commonly has 10000 iterations, so that it takes around 17 minutes to train the NP-ODE. For the comparison, the training process of the original NPs takes around 5 minutes for 10000 iterations, and the training process of the GP takes around 0.05 seconds.

It is worth mention that, as the surrogate model, the training process is most time-consuming and is often taken offline. As long as the model is ready to use, it is no need to repeat the training process. The computational cost of the NP-ODE in predicting 20 testing data points is 0.2 seconds, which is comparable to the original NPs and the GP. More importantly, comparing the time efficiency of surrogate models to the FEA method is more practically meaningful. Compared with the FEA method, the trained NP-ODE significantly reduces the computational cost of predicting results on new inputs from hours to seconds.

## 6. Conclusion

Despite the strength and accuracy of the FEA method, its applications are hindered by the significant computational cost and lack of ability in uncertainty quantification. Since uncertainties inevitably exist in real systems, uncertainty quantification is essential in system modeling. The Monte Carlo method is a standard approach to conduct uncertainty quantification which generates FEA simulations repetitively with various parameters. However, the Monte Carlo method suffers from significantly growing computational time when repeating simulations. The existing surrogates built for uncertainty quantification of FEA are mainly based on the Gaussian process and its variants. Although it reduces computational cost compared with the FEA method, it suffers from lacking interpretability and unable to handle large volume and high-dimensional data.

This paper proposes an FEA-informed stochastic surrogate NP-ODE to model the FEA simulations as well as evaluating the uncertainties of the output. In the NP-ODE, the basic structure follows the original NPs which enables the model to generate distributions of output for uncertainty quantification, and the Neural-ODE is incorporated as the decoder, which improves the model's ability in solving systems governed by differential equations. In the case study, to validate and compare the performances of our proposed NP-ODE and benchmark models, we select MAPE and 95% confidence interval to evaluate the predictive accuracy and uncertainty quantification. From the comparisons of the results, the advantages of our proposed NP-ODE can be summarized into several aspects: a) compared with the Gaussian process and its variants, our proposed NP-ODE shows a better ability in exploring limited training samples and has a robust performance in both predictive error and uncertainty quantification when decreasing the training samples; b) compared with original NPs, the incorporation of Neural-ODE reduces the number of model parameters and enables our proposed NP-ODE to better model FEA simulations; (c) the proposed NP-ODE solves differential equations in its decoders, so it is more physically close to the mechanism of original FEA.

There are also several limitations of the proposed NP-ODE. Firstly, incorporating Neural-ODE might introduce extra computational cost if we select a small value of step size $\Delta D$. In our experiment, the influence is mitigated by tuning the value of $\Delta D$ to find a balance between accuracy and efficiency. Secondly, given the accurate results of the simulation study and case study, the assumption of Euler's method is to fix the value of $\Delta D$ in solving NP-ODE. Other numerical methods with adaptive step size can be the alternatives to solve NP-ODE.

For future extensions, the proposed NP-ODE may provide a prototype about how to account for the system uncertainties and design stochastic surrogates for FEA simulations. Following this prototype, it will be promising to design stochastic surrogates for systems governed by PDEs by developing a new decoder accordingly.

**Acknowledgment**

This research was financially supported by the US National Science Foundation under Grant CMMI-1855651. The computational resource used in this work is provided by the advanced research computing at Virginia Polytechnic Institute and State University.

**Data and Code Availability**

The dataset and codes for this paper will be available upon publication.

## Appendix I

Table 4. Math Notations

| | |
|---|---|
| $(x_i, y_i)$ | Data points with scalar input and scalar output. |
| $(\mathbf{x}_i, y_i)$ | Data points with vector input and scalar output. |
| $(\mathbf{x}_i, \mathbf{y}_i)$ | Data points with vector input and vector output. |
| $D$ | Equivalent model depth in Neural-ODE. |
| $\Delta D$ | Step size in Euler's method. |

| Symbol | Description |
|---|---|
| $(\mathbf{d}_1, \ldots, \mathbf{d}_n)$ | Deterministic representation of observed data points. |
| $\mathbf{W}_{DE}, \mathbf{b}_{DE}$ | Weight matrix and bias in the FC layer of the deterministic encoder. |
| $\mathbf{d}_C$ | Aggregated deterministic representation from the attention module. |
| $(\mathbf{s}_1, \ldots, \mathbf{s}_n)$ | Stochastic representation of observed data points. |
| $\mathbf{W}_{SE}, \mathbf{b}_{SE}$ | Weight matrix and bias in the FC layer of the stochastic encoder. |
| $\mathbf{s}_C$ | Mean aggregated stochastic representation. |
| $\mathbf{z}$ | Latent representation accounting for prediction uncertainties. |
| $\bar{y}_{n+1}$ | The predicted mean for scalar output. |
| $\sigma_{n+1}$ | The predicted standard deviation for scalar output. |
| $\hat{y}_{n+1}$ | Sampled scalar predictive results from output distribution. |
| $\bar{\mathbf{y}}_{n+1}$ | The predicted mean for vector output. |
| $\bar{\mathbf{y}}_{n+1}^i$ | $i_{th}$ element in the predicted mean vector. |
| $\boldsymbol{\sigma}_{n+1}$ | The predicted standard deviation for vector output. |
| $\hat{\mathbf{y}}_{n+1}$ | Sampled vector predictive results from output distribution. |
| $f(\mathbf{x})$ | The underlying functional relationship of $(\mathbf{x}_i, y_i)$. |
| $\mathbf{f}(\mathbf{x})$ | The underlying functional relationship of $(\mathbf{x}_i, \mathbf{y}_i)$. |
| $\hat{f}_{NN}(x)$ | Approximated mapping given by NN. |
| $\hat{f}_{NODE}(D)$ | Discrete transformation given by Neural-ODE with scalar as output. |
| $g_{NODE}(D)$ | First-order derivative of $\hat{f}_{NODE}(D)$. |
| $\hat{\mathbf{f}}_{NODE}(D)$ | Discrete transformation given by Neural-ODE with vector as output. |
| $h_i(.)$ | The function of the FC layer with vector as input and scalar as output. |
| $\mathbf{g}_{NODE}(D)$ | First-order derivative of $\hat{\mathbf{f}}_{NODE}(D)$. |
| $p(\hat{y}_{n+1}|\mathbf{x}_{n+1}, (\mathbf{x}_{1:n}, \mathbf{y}_{1:n}))$ | Posterior distribution of predictions given observed data points. |
| $q(\mathbf{z}|\mathbf{s}_C)$ | Prior distribution on $\mathbf{z}$ given by stochastic encoder. |
| $p(\hat{y}_{n+1}|\mathbf{x}_{n+1}, \mathbf{d}_C, \mathbf{z})$ | Likelihood of predictions given by decoder. |
| $\mathcal{D}_T$ | The dataset containing target (unobserved) data points. |
| $\mathcal{D}_C$ | The dataset containing context (observed) data points. |
| $x_T$ | The dataset containing the input of target data points. |

**Appendix II**

The detailed derivation of equation (4) is given below, in which $(\mathbf{x}_{1:n}, \mathbf{y}_{1:n})$ are observed (context) data points, $(\mathbf{x}_{n+1:n+T}, \mathbf{y}_{n+1:n+T})$ are unobserved (target) data points, $\mathbf{d}_C$ is the deterministic representation of context data points, $\mathbf{z}$ is the latent representation of target data points.

$$\log p(\mathbf{y}_{n+1:n+T}|\mathbf{x}_{n+1:n+T}, (\mathbf{x}_{1:n}, \mathbf{y}_{1:n}))$$
$$\geq \mathbb{E}_{q(\mathbf{z}|(\mathbf{x}_{1:n+T}, \mathbf{y}_{1:n+T}))} \left[ \sum_{i=1}^{T} \log p(\mathbf{y}_{n+i}, \mathbf{z}|\mathbf{x}_{n+i}, \mathbf{d}_C) - \log q(\mathbf{z}|\mathbf{x}_{1:n+T}, \mathbf{y}_{1:n+T}) \right]$$
$$= \mathbb{E}_{q(\mathbf{z}|(\mathbf{x}_{1:n+T}, \mathbf{y}_{1:n+T}))} \left[ \sum_{i=1}^{T} \log p(\mathbf{y}_{n+i}|\mathbf{x}_{n+i}, \mathbf{d}_C, \mathbf{z}) + \log \frac{p(\mathbf{z}|\mathbf{x}_{1:n}, \mathbf{y}_{1:n})}{q(\mathbf{z}|\mathbf{x}_{1:n+T}, \mathbf{y}_{1:n+T})} \right]$$

(4*)

Considering the $p(\mathbf{z}|\mathbf{x}_{1:n}, y_{1:n})$ is intractable, it is approximated by variational posterior $q(\mathbf{z}|\mathbf{x}_{1:n}, \mathbf{y}_{1:n})$, so that the derivation is followed by:

$$= \mathbb{E}_{q(\mathbf{z}|(\mathbf{x}_{1:n+T}, \mathbf{y}_{1:n+T}))} \left[ \sum_{i=1}^{T} \log p(\mathbf{y}_{n+i}|\mathbf{x}_{n+i}, \mathbf{d}_C, \mathbf{z}) + \log \frac{q(\mathbf{z}|\mathbf{x}_{1:n}, \mathbf{y}_{1:n})}{q(\mathbf{z}|\mathbf{x}_{1:n+T}, \mathbf{y}_{1:n+T})} \right]$$
$$= \mathbb{E}_{q(\mathbf{z}|(\mathbf{x}_{1:n+T}, \mathbf{y}_{1:n+T}))} \left[ \sum_{i=1}^{T} \log p(\mathbf{y}_{n+i}|\mathbf{x}_{n+i}, \mathbf{d}_C, \mathbf{z}) \right] - D_{KL}(q(\mathbf{z}|\mathbf{x}_{1:n+T}, \mathbf{y}_{1:n+T}) \| q(\mathbf{z}|\mathbf{x}_{1:n}, \mathbf{y}_{1:n}))$$
$$= \mathbb{E}_{q(\mathbf{z}|\mathbf{s}_T)} \left[ \sum_{i=1}^{T} \log p(\mathbf{y}_{n+i}|\mathbf{x}_{n+i}, \mathbf{d}_C, \mathbf{z}) \right] - D_{KL}(q(\mathbf{z}|\mathbf{s}_T) \| q(\mathbf{z}|\mathbf{s}_C))$$

(4*)

In the last step, the target and context data points are reparameterized by the stochastic decoder, which is denoted by $\mathbf{s}_T$ and $\mathbf{s}_C$ respectively (Kingma and Welling, 2014).

**Appendix III**

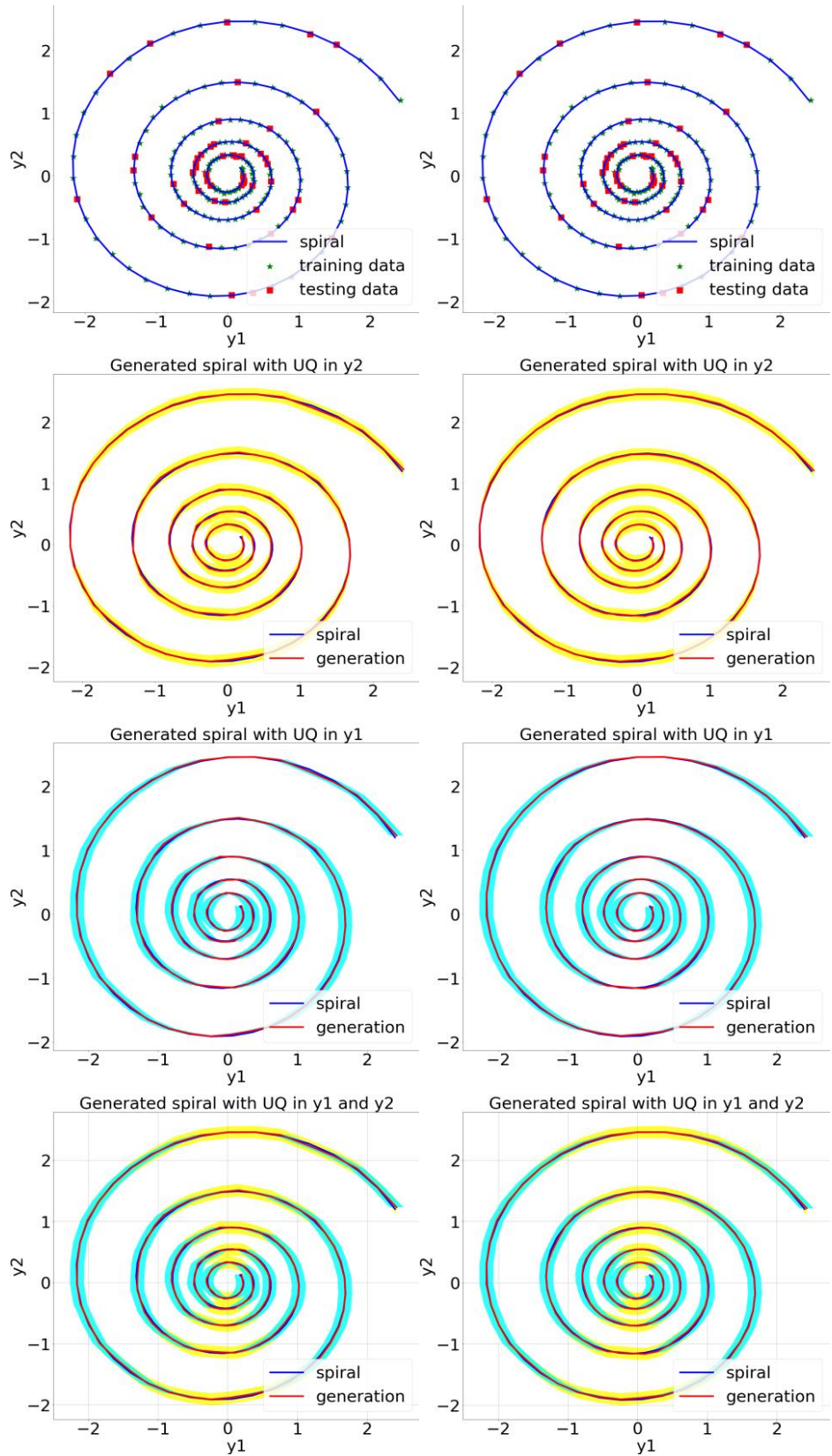

Figure 10. UQ Comparison with noise $\sigma = 0.01$; **Left:** result from NP-ODE; **Right**: result from NPs

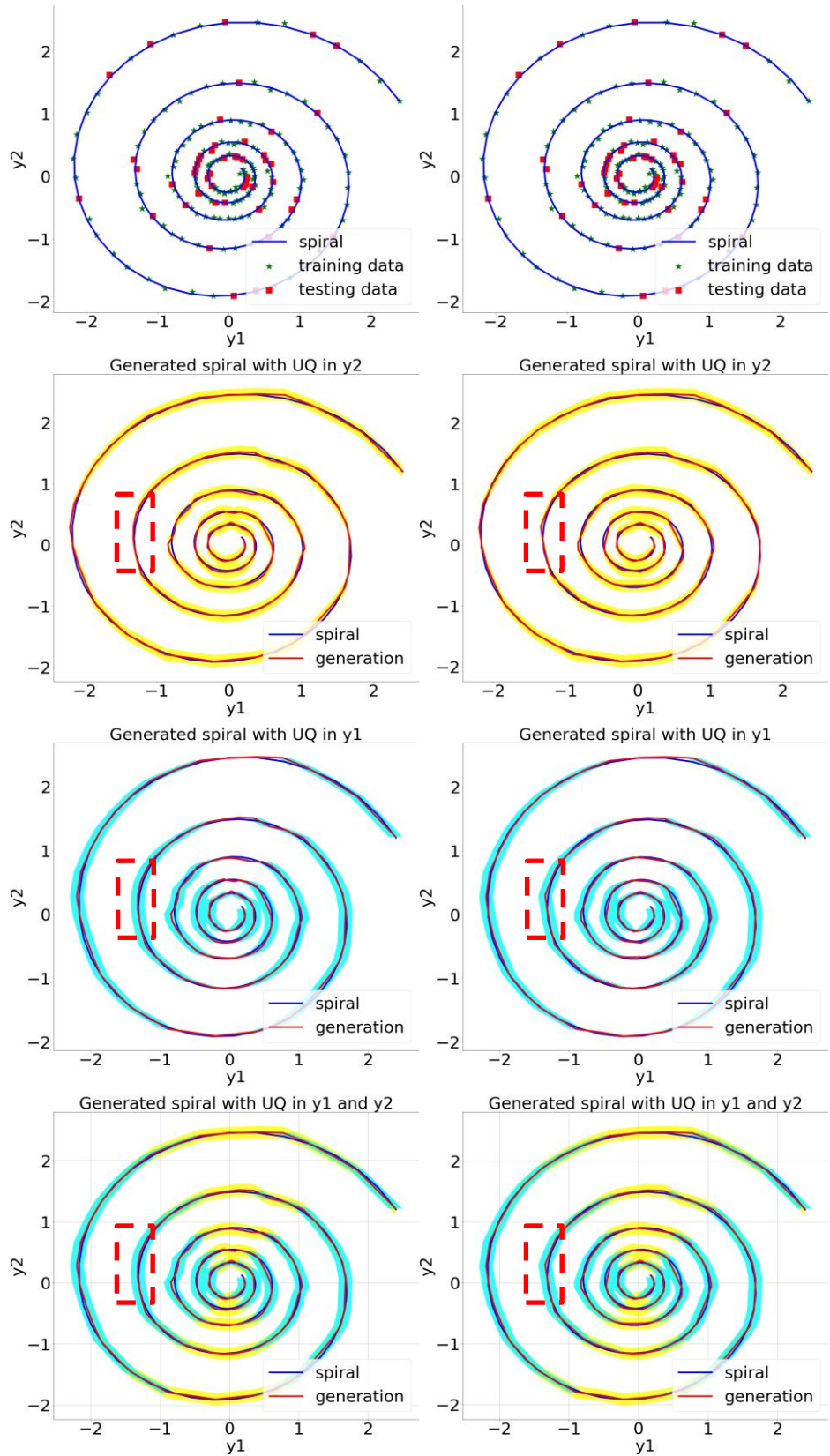

Figure 11. UQ Comparison with noise $\sigma = 0.02$; **Left:** result from NP-ODE; **Right**: result from NPs